\documentclass{article}
\usepackage{ijcai15}

\usepackage{times}

\usepackage{latexsym}
\usepackage{multirow}
\usepackage{graphicx}
\usepackage{amsmath}
\usepackage{mathrsfs}
\usepackage{amssymb}
\usepackage{amsfonts}
\usepackage{color}

\newcommand{\namecite}[1]{\citeauthor{#1}~\shortcite{#1}}

\title{Local Translation Prediction with Global Sentence Representation}
\author{ Jiajun Zhang\\
National Laboratory of Pattern Recognition, CASIA, Beijing, P.R. China\\
 jjzhang@nlpr.ia.ac.cn\\
}

\begin{document}

\maketitle

\begin{abstract}
Statistical machine translation models have made great progress in improving the translation quality. However, the existing models predict the target translation with only the source- and target-side local context information. In practice, distinguishing good translations from bad ones does not only depend on the local features, but also rely on the global sentence-level information. In this paper, we explore the source-side global sentence-level features for target-side local translation prediction. We propose a novel bilingually-constrained chunk-based convolutional neural network to learn sentence semantic representations. With the sentence-level feature representation, we further design a feed-forward neural network to better predict translations using both local and global information. The large-scale experiments show that our method can obtain substantial improvements in translation quality over the strong baseline: the hierarchical phrase-based translation model augmented with the neural network joint model.
\end{abstract}

\section{Introduction}
In the recent years, statistical machine translation (SMT) models, such as phrase-based models \cite{koehn2007moses}, hierarchical phrase-based models \cite{chiang2007hierarchical}, and linguistically syntax-based models \cite{liu2006tree,galley2006tree}, have achieved great progress in improving the translation performance. In these translation models, the target sentence is generated by compositing several local translations with reordering models or synchronous grammars, and the local translations are rendered with the help of the source- and target-side local context information. In most cases, the translation of source language words can be determined with local context features. However, there are many cases in which the target translation does not only depend on the local context, but also rely on the global sentence-level information.

Take the Chinese sentence and its English translation in Figure 1 as an example. For the Chinese word in red color, {\em molecule} is the most possible translation. Even with the help of the local context information, it cannot figure out the correct translation. Given the sentence-level semantics talking about {\em preventing illegal immigrants and unsafe people from entering Japan}, we can make sure that the best target translation for the Chinese word in red color should be {\em people}. Obviously, the global sentence-level semantic information plays an important role in local translation prediction.

\begin{figure}
\centering
\includegraphics[scale=0.36]{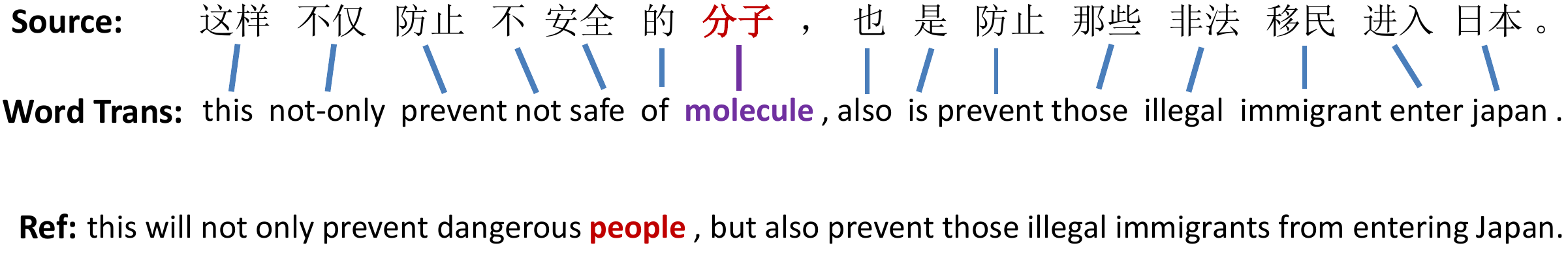}
\caption{An example for Chinese to English translation in which the translation of the Chinese word in red needs the global sentence information of the Chinese sentence.}
\label{Fig.1}
\end{figure}

Two questions arise: 1) how to represent the global sentence-level semantics? 2) how to make full use of the sentence semantic representation in statistical machine translation models?

\begin{figure*}
\centering
\includegraphics[scale=0.55]{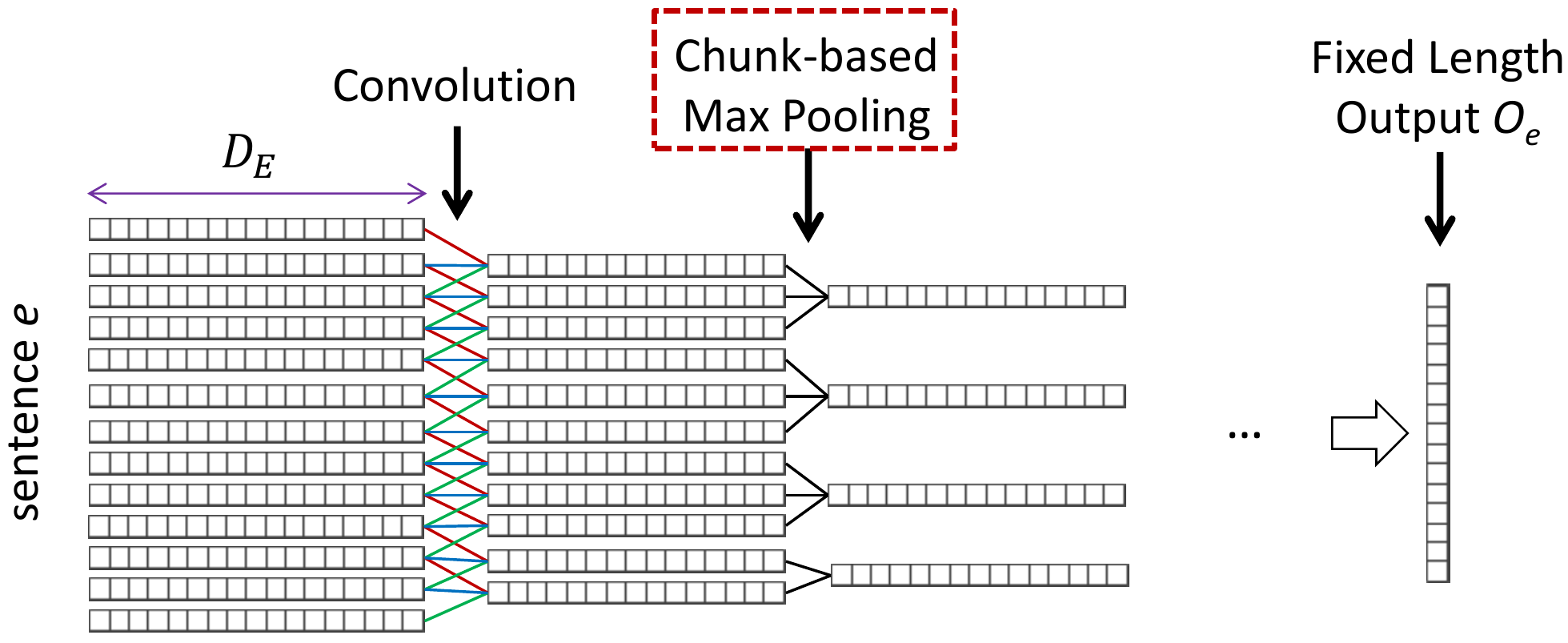}
\caption{Chunk-based convolutional neural network architecture for modelling sentence representation.}
\label{Fig.2}
\end{figure*}

For the first problem, the neural network methods are proposed recently to learn sentence representations. These methods include recurrent neural networks \cite{kalchbrenner2013recurrent,sutskever2014sequence,bahdanau2014neural}, recursive neural networks \cite{socher2011parsing,socher2013parsing,socher2013recursive}, sentence to vector \cite{le2014distributed} and convolutional neural networks \cite{kim2014convolutional,kalchbrenner2014convolutional,hu2014convolutional,zeng2014relation}. It should be noted that most of the above approaches learn the distributed sentence representations for specific tasks, such as classification, sequence labelling, and structure prediction. The semantic meaning of the sentence is not fully contained by the sentence representation. In this paper, we focus on learning the sentence semantic representation. Although we have no gold sentence semantic representation available for supervised learning, we have a large amount of parallel sentence pairs which share the same semantic meaning. Accordingly, sentence translation equivalents can supervise each other to learn their semantic representations. Furthermore, we design the chunk-based convolutional neural network in order to well handle the sentence length variation and retain as much information as possible at the same time. In this network, we just need to decide how many chunks we want to segment a sentence. Therefore, combining the two ideas together, we propose the bilingually-constrained chunk-based convolutional neural network (BCCNN) for sentence semantic representation.

For the second problem, we incorporate the sentence semantic representation during the decoding process to better generate the target translation. Following the idea in \cite{devlin2014fast}, we design a feed-forward neural network which takes the learnt sentence semantic representation as the new input feature to predict the conditional probability of the target word given both the local and global information. As an informative feature, this conditional probability is integrated into the log-linear model of the hierarchical phrase-based translation system \cite{chiang2007hierarchical}.

In this paper, we make the following contributions:
\begin{itemize}
\item Our idea of the bilingually-constrained method circumvents the problem of lacking gold labelled data and provides a good way to learn sentence semantic representations.

\item To deal with the variable length of the sentences and meanwhile retain as much semantics as possible, we propose the chunk-based convolutional neural network in which we can choose the number of the chunks.

\item When integrating the sentence semantic representation into the decoding process, we can achieve significant improvements in translation quality over a strong baseline.

\end{itemize}

\section{Sentence Semantic Representation}
Convolutional neural network (CNN) consisting of the convolution and pooling layers provides a standard architecture \cite{collobert2011natural} which maps variable-length sentences into fixed-length distributed vectors. This section starts with introducing a new variant of the standard CNN called chunk-based CNN in order to keep more semantics of the sentence.

\subsection{Chunk-based Convolutional Neural Network}
The model architecture is illustrated in Figure 2. The data flow is similar to the standard CNN: the model takes as input the sequence of word embeddings in a sentence, summarizes the sentence meaning by convolving the sliding window and pooling the saliency through the sentence, and yields the fixed-length distributed vector with other layers, such as dropout layer, more convolution and pooling layers, linear and non-linear layers.

Specifically, assuming we are equipped with a word embedding matrix ${\mathbb{L}} \in {\mathbb{R}}^{k \times |V|}$ {\footnote{$k$ is the embedding dimension and $|V|$ is the vocabulary size.}} trained on the large-scale monolingual data using unsupervised algorithm (e.g. word2vec \cite{mikolov2013distributed}). Given a sentence $w_1w_2 \cdots w_n$, each word $w_i$ is first projected into a vector $X_i$ through the word embedding matrix. Then, we concatenate all the vectors in order to form the input of the model:
\begin{equation}
X = [X_1, X_2, \cdots, X_n]
\end{equation}

{\bf Convolution Layer} involves a number of {\em filters} $W \in {\mathbb{R}}^{h \times k}$ which summarize the information of $h$-word window and produce a new feature. For the window of $h$ words $X_{i:i+h-1}$, a {\em filter} $F_l$ ($1 \le l \le L$, and $L$ denotes the number of filters) generates the feature $y^l_i$ as follows:
\begin{equation}
y^l_i = \sigma (W \cdot X_{i:i+h-1} + b)
\end{equation}
Where $\sigma$ is a non-linear activation function (e.g. Relu or Sigmoid), and $b$ is a bias term. When a {\em filter} traverses each window in the sentence from $X_{1:h-1}$ to $X_{n-h+1:n}$, we get the output of the feature map corresponding to the {\em filter} $F_l$:
\begin{equation}
y^l = [y^l_1, y^l_2, \cdots, y^l_{n-h+1}]
\end{equation}
Here, $y^l \in {\mathbb{R}}^{n-h+1}$. It should be noted that the sentences differ with each other in length $n$ (from several words to more than 100 words), and then $y^l$ has different dimensions for different sentences. It becomes a key question how to transform the variable-length vector $y^l$ into a fixed-length vector.

\begin{figure*}
\centering
\includegraphics[scale=0.55]{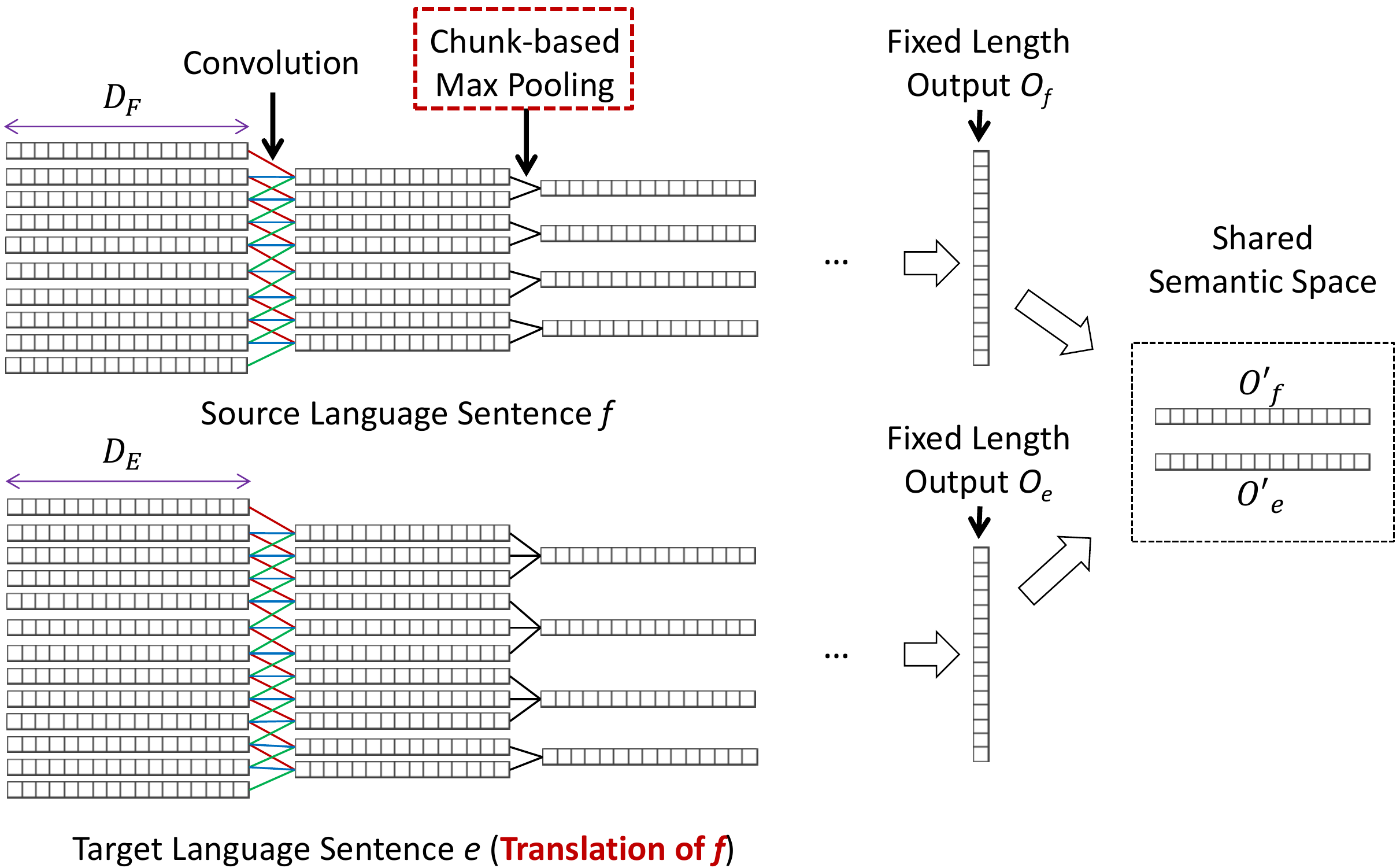}
\caption{Bilingually-constrained Chunk-based convolutional neural network architecture for learning sentence semantic representation.}
\label{Fig.3}
\end{figure*}

{\bf Pooling Layer} is designed to perform this task. At most cases, we apply a standard {\em max-over-time} pooling operation \cite{collobert2011natural,kim2014convolutional} over $y^l$ and choose the maximum value $\hat{y}^l=max\{y^l\}$ as the most important feature of the filter $F_l$. This idea is simple and easy to implement. However, its disadvantage lies in two-fold: 1) most of the information in the sentence is lost, and 2) the word order information is missing. \namecite{kalchbrenner2014convolutional} proposed to take the top-$K$ maximum values over $y^l$ to keep more information, but the word order information is still missing. \namecite{hu2014convolutional} designed a max-pooling over every two-unit, but they require the fixed-length inputs. In this paper, we design a simple but effective {\bf chunk-based max-pooling} so as to retain more semantics of the sentence and keep the order information as well. 

Given the predefined number of chunks $C$ (e.g. $C=4$ in Figure 2), we first divide evenly $y^l$ into $C$ segments, and then takes the maximum value from each segment. Note that $n-h+1$ does not have to be divisible by $C$ and the last chunk can have the size of the modulus. Then, we can transform the variable-length vector $y^l$ into a fixed-length vector with $C$ values:
\begin{equation}
\begin{split}
y^{lC} & = chunkMax\{y^l\} \\
  & = chunkMax\{[y^l_1, y^l_2, \cdots, y^l_{n-h+1}]\} \\
  & = [y^l_{c_1}, y^l_{c_2}, \cdots, y^l_{C}]
\end{split}
\end{equation}
If there are $F_L$ {\em filters}, the output of the pooling layer will be a vector in $F_L \times C$ dimension. In our model, the pooling layer is followed by a dropout layer, two fully connected linear layers with non-linear activation Relu. Finally, we can obtain a fixed-length output vector for each sentence.

This idea of chunk-based CNN is inspired by the inherent structure of a sentence. From the perspective of shallow structures, a sentence is organized by subject-verb-object (English and Chinese word order). Then, we can assign $C$ with a small number (e.g 2 or 3) to summarize this kind of semantic information. From the perspective of deep structures, a sentence can be described as a sequence of noun phrase (NP), verb phrase (VP), adjective phrase (ADJP), prepositional phrase (PP) and so on. Accordingly, we can set $C$ to be a relatively large number to capture these kinds of information. Therefore, the chunk-based CNN can retain more semantics of the sentence and maintain as well the sentence structure to some extent. 

\subsection{Bilingually-constrained Chunk-based CNN}
The convolutional neural network is usually tuned to optimize an objective function for a specific task, such as sentiment and relation classification \cite{kim2014convolutional,zeng2014relation}. The result sentence representation is class sensitive but does not encode adequate semantic meaning of the sentence. Since our goal is to learn sentence semantic representations, we need to find a well-defined objective function.

However, there are no gold sentence semantic representations available in the real world. Fortunately, we know the fact that if two sentences share the same meaning, their semantic representations should be identical. As we know in machine translation that, there are lots of parallel sentence pairs for different languages, such as Chinese-English, Arabic-English. Thus, we can make an inference from this fact that if a model can learn the same representation for any parallel sentence pair sharing the same meaning, the learnt representation must encode the semantics of the sentences and the corresponding model is our desire.  Inspired by the work on word and phrase embeddings using a bilingual method \cite{zou2013bilingual,zhang2014brae}, we propose the Bilingually-constrained Chunk-based CNN (BCCNN), whose basic goal is to minimize the semantic distance between the sentences and their translations.

As illustrated in Figure 3, given a source language sentence $f$ and its translation $e$, the chunk-based CNN can generate respectively the fixed-length output vectors $O_f$ and $O_e$. Then, $O_f$ and $O_e$ are projected into a shared semantic space, and become $O'_f$ and $O'_e$ (e.g. $O'_f = \sigma(W^t \cdot O_f + b^t)$, where $W^t$ denotes transformation matrix). In the shared space, two representation vectors can calculate their distance easily with dot-product. The basic objective function is to minimize the distance $dis(f,e;\Theta)=dis(O'_f,O'_e)$ between $O'_f$ and $O'_e$.

We know that a good model should not only make the representations of translation equivalents as similar as possible, but also should enforce the representations of non-translation pairs as different as possible. Therefore, our objective function is also designed to maximize the distance $dis(f,e^*;\Theta)$ if $(f,e^*)$ is a non-translation pair. Then, we design our objective function to be a max-margin loss:
\begin{equation}
j(f,e,e^*;\Theta) = max(0,1+dis(f,e^*;\Theta)-dis(f,e;\Theta))
\end{equation}
Here, $\Theta$ includes all the parameters of the bilingually-constrained chunk-based CNN. For any translation equivalent $(f,e)$, we can choose randomly a sentence $e^*$ ($e^* \neq e$) from the target language monolingual data and obtain the non-translation pair $(f,e^*)$. The finally objective function over the large-scale parallel sentence pairs $(F,E)$ of size $N$ will be:
\begin{equation}
J(F,E;\Theta) = \frac{1}{N} \sum\limits_{(f,f)\in (F,E)} j(f,e,e^*;\Theta) + \frac{\lambda}{2}{||\Theta||}^2
\end{equation}

\section{Integrating Sentence Semantic Representations in Translation Models}
With the trained chunk-based convolutional neural network, we can obtain the semantic representation for any sentence. In this section, we introduce how to make full use of the sentence semantic representations in statistical machine translation models. 

\subsection{Sentence Representation for Translation Probability Estimation}
Formally, given a source sentence $s$, machine translation aims to find from the search space $T$ the best target translation hypothesis $t$ which has the highest conditional probability $p(t|s)$. If we focus on each target word, the conditional probability can be decomposed as follows:
\begin{equation}
p(t|s) = \prod_{i=1}^{|t|}p(t_i|t_1,t_2,\cdots,t_{i-1},s)
\end{equation}

\namecite{devlin2014fast} approximated the target single word probability $p(t_i|t_1,t_2,\cdots,t_{i-1},s)$ following the target $n$-gram language model and using the source-side local context, called joint model:
\begin{equation}
p(t|s) \approx \prod_{i=1}^{|t|}p(t_i|t_{i-n+1},\cdots,t_{i-1},\mathcal{S}_i)
\end{equation}
In which $\mathcal{S}_i$ includes the source-side local context associated with the current target word $t_i$. We know that machine translation models, such as hierarchical phrase-based model, generate the target hypothesis with translation rules {\footnote{Word alignment information in the translation rules are retained during decoding.}} from which we can find the source word $s_{a_i}$ that is aligned to $t_i$ ($s_{a_i}$ and $t_i$ are translations with each other). \namecite{devlin2014fast} take the $m$-word source-side local context centered at $s_{a_i}$:
\begin{equation}
\mathcal{S}_i = s_{a_i-\frac{m-1}{2}}, \cdots, s_{a_i}, s_{a_i+\frac{m-1}{2}}
\end{equation}

However, just as we discussed in the Introduction section that besides the local context, the global sentence semantic information plays an indispensable role in accurate translation prediction. Therefore, we augment Equation 8 with the global sentence semantics:
\begin{equation}
p(t|s) \approx \prod_{i=1}^{|t|}p(t_i|t_{i-n+1},\cdots,t_{i-1},\mathcal{S}_i, {\bf \color{blue} s})
\end{equation}
In this way, translating every target word during decoding is aware of the source-side global sentence-level information.

In our experiments, following \cite{devlin2014fast} we use $n=4$ and $m=11$. It is easy to see that the data sparsity will become a serious problem if we employ the traditional method to perform the probability estimation. Therefore, we resort to the neural networks that perform the probability estimation in a distributed continuous space.

\begin{figure}
\centering
\includegraphics[scale=0.45]{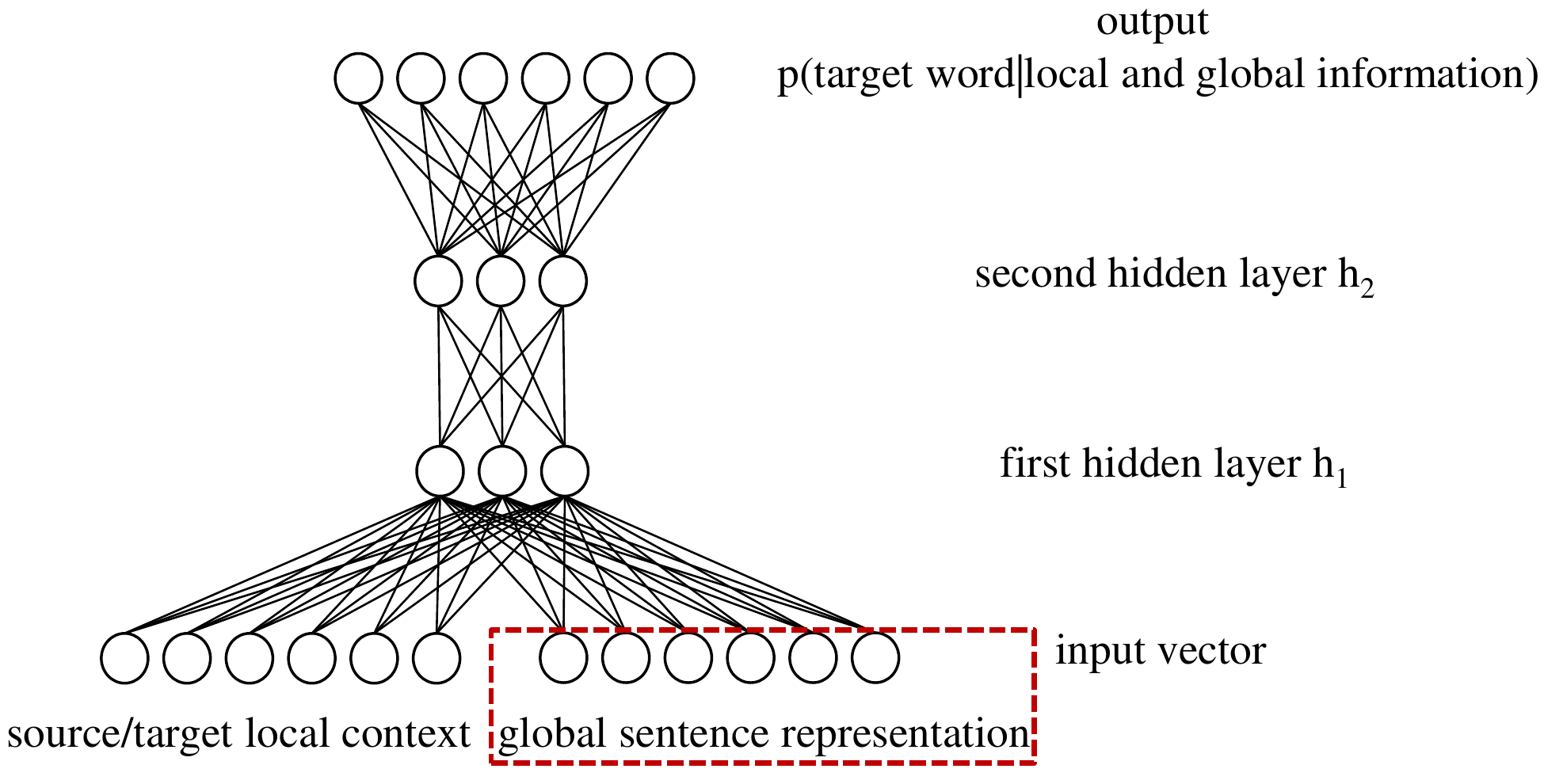}
\caption{Local translation prediction with both the local context and the global sentence semantic representations.}
\label{Fig.4}
\end{figure}

The neural network architecture shown in Figure 4 is almost identical to the original feed-forward neural network described in \cite{bengio2003neural,vaswani2013nn,devlin2014fast}. It consists of two hidden layers besides the input and output layer. The input includes two parts: 1) the local $m+n-1$ context vector used in \cite{devlin2014fast} ($n-1$ target history words and $m$ source-side context words), where each word is mapped to a 192-dimensional vector; 2) the global 192-dimensional sentence semantic representation vector obtained with the learnt chunk-based CNN.

Through two 512-dimensional hidden layers with rectified linear activation function (Relu, $\sigma(x)=max(0,x)$), we apply the softmax function in the output layer to calculate the probability for each target word in the vocabulary. Following \cite{devlin2014fast}, input vocabulary contains 16,000 source words and 16,000 target words. The output vocabulary contains 32,000 target words.

This feed-forward neural network will be trained to maximize the log-likelihood over the target side of the bilingual training data for machine translation.

\subsection{Decoding with Neural Network Probability}
To calculate the target word conditional probability with neural network, we need the information of $n-1$ target history words, $m$ source context words and the whole source sentence, which are all easy to obtain during SMT decoding. Thus, the neural network probability can be integrated into any SMT model. In this paper, the hierarchical phrase-based model (HPB) \cite{chiang2007hierarchical} is employed.

The HPB model translates the source sentences using synchronous context free grammars (SCFG). The SCFG translation rules are in the form of $X \rightarrow (\alpha, \gamma, \sim)$, where $X$ is the non-terminal symbol, $\alpha$ and $\gamma$ are sequences of lexical items and non-terminals in the source and target side respectively, and $\sim$ indicates one-to-one correspondence between non-terminals in $\alpha$ and $\gamma$ in the standard version. Here, in order to retrieve easily the central source word aligned to each target word during decoding, we also retain the alignments between source- and target-side lexical terms in the rules. That is, $\sim$ contains all the correspondences between terminal or non-terminals in $\alpha$ and $\gamma$.

The HPB SMT adopts a log-linear model to search for the best translation candidate. The powerful features in the log-linear model includes: 1) forward and backward rule probability, 2) forward and backward lexical probability, 3) a 5-gram language model, 4) rule counts and translation length penalty, and 5) a glue rule reordering model. The conditional probability $p(t_i|t_{i-n+1},\cdots,t_{i-1},\mathcal{S}_i,s)$ calculated in the previous section will serve as the sixth kind of the informative feature to be integrated in the log-linear model.

\section{Experiments}
Before elaborating the experimental results, we first introduce the details of neural network training and the experimental settings.

\subsection{Neural Network Training Details}
For the bilingually-constrained chunk-based CNN, the initial 192-dimensional word embeddings are trained with word2vec \cite{mikolov2013distributed}: the embedding for English words is learnt on $\sim$1.1B data, while that for Chinese words is learnt on $\sim$0.7B data. We set the context window $h=3$ for convolution. We will test multiple settings of the chunk number ($C=1,2,4,8$) to see which one performs best. We apply $L=100$ filters. The two fully connected linear layers both contain 192 neurons. The dropout ratio in the dropout layer is set 0.5 to prevent overfitting. Dot-product is employed to calculate the distance between source and target sentence representations in the shared semantic space. The standard back-propagation and stochastic gradient descent (SGD) algorithm is utilized to optimize this network.

For the feed-forward neural network, we also apply the SGD algorithm. A key issue of this neural network is that the computation in the softmax layer is too time consuming since normalization is required over the entire huge vocabulary. Inspired by \cite{vaswani2013nn}, we adopt the Noisy Contrastive Estimation (NCE) \cite{gutmann2010noise} to avoid the normalization in the output layer.

\subsection{Experimental Setup}

\begin{table} \small
\begin{center}
\begin{tabular}{c|r|r}
\hline
Data & Chinese Sent. Num. & English Sent. Num. \\
\hline
bilingual data & 2,086,731 & 2,086,731 \\
\hline
Xinhua News & $\sim$ & 10,912,683 \\
\hline
NIST03 & 919 & 919$\times$4 \\
\hline
NIST05 & 1,082 & 1,082$\times$4 \\
\hline
NIST06 & 1,000 & 1,000$\times$4 \\
\hline
NIST08 & 691 & 691$\times$4 \\
\hline
\end{tabular}
\caption{\label{Table 1} Data statistics of the SMT experiment.}
\end{center}
\end{table}

The SMT evaluation is conducted on Chinese-to-English translation. The bilingual training data {\footnote{LDC2000T50, LDC2002L27, LDC2002T01, LDC2003E07, LDC2003E14, LDC2003T17, LDC2004T07, LDC2005T06, LDC2005T10 and LDC2005T34.}} from LDC contains about 2.1 million sentence pairs. This bilingual data is also utilized to train the two neural networks. The 5-gram language model is trained on the English part of the bilingual training data and the Xinhua portion of the English Gigaword corpus. NIST MT03 is used as the tuning data. MT05, MT06 and MT08 (news data) are used as the test data. Table 1 shows the detailed data statistics. Case-insensitive BLEU is employed as the evaluation metric. The statistical significance test is performed with the pairwise re-sampling approach \cite{koehn2004sigtest}.

\subsection{Experimental Results}
To have a comprehensive understanding about the capacity of our proposed model, we compare our method with several baselines. The information of different systems is detailed as follows:
\begin{itemize}
\item {\bf HPB}: the hierarchical phrase-based translation system.

\item {\bf +NNJM}: HPB system incorporating feed-forward neural network joint model in which the probability is predicted with $3$ target history words and $11$ source-side local context words.

\item {\bf +AVE\_SENT}: it is similar to +NNJM, but the neural network probability is augmented with source-side global sentence representation which is obtained by averaging all the word embeddings in the sentence.

\item {\bf +BCCNN}: it is similar to +AVE\_SENT. Instead of average embedding, the sentence semantic representation is learnt using the bilingually-constrained chunk-based convolutional neural networks.
\end{itemize}

\begin{table}\small
\begin{center}
\begin{tabular}{l|c|c|c|c}
\hline \bf System & \bf MT03 & \bf MT05 & \bf MT06 & \bf MT08 \\ \hline
HPB & 35.98 & 34.66 & 35.25 & 27.80  \\
\hline
\hline
+NNJM & 36.93 & ${35.55}^+$ & 35.77 & ${28.64}^+$  \\ 
+AVE\_SENT	& 37.16 & ${35.88}^+$ & ${36.07}^+$ & ${29.19}^+$  \\ 
+BCCNN-1 & 37.32 & ${36.06}^+$ & ${36.42}^+$ & ${29.35}^{+*}$  \\
+BCCNN-2 & 37.75 & ${36.24}^+$ & ${36.65}^{+*}$ & ${29.97}^{+*}$  \\
+BCCNN-4 & ${\bf 37.98}$ & ${36.22}^+$ & ${\bf 36.78}^{+*}$ & ${\bf 30.02}^{+*}$  \\
+BCCNN-8 & 37.64 & ${\bf 36.29}^{+*}$ & ${36.49}^+$ & ${29.98}^{+*}$  \\
	\hline
\end{tabular}
\caption{\label{Table 2} Experimental results of different translation systems. Significance test is performed on the test sets. "$+$" means that the model significantly outperforms the baseline HPB with $p < 0.01$. "$*$" indicates that the model is significantly better than +NNJM with $p < 0.01$. "+BCNN-4" denotes that this model adopts 4 chunks in the pooling layer.}
\end{center}
\end{table}

Table 2 gives the detailed results. First, let us look at the performance of the neural network joint model using the local contexts (+NNJM). Compared to the hierarchical phrase-based model HPB, this model performs significantly better on test set MT05 and MT08. The biggest improvement can be up to 0.95 BLEU score. \namecite{devlin2014fast} has reported that this model can outperform HPB by more than 1.0 BLEU score on Chinese-to-English translation. Although our improvement is not so promising, we demonstrate that it is much helpful to apply the neural network joint model using source- and target-side local contexts.

When the model +NNJM is augmented with the global sentence representations, we can obtain more gains (see last 5 lines in Table 2). Specifically, if the sentence representation is generated by just averaging all the word embeddings, it can get slight improvements over the model +NNJM. However, +AVE\_SENT cannot perform significantly better than +NNJM. These results indicate that the sentence representation is beneficial to improving the translation quality. Due to the lack of adequate semantics, it does not lead to great improvements.

As we can see that if the sentence representation is learnt by the bilingually-constrained chunk-based CNN (BCCNN), the models can achieve much more BLEU score improvements no matter how many chunks we adopt. Note that using only one chunk is equivalent to the {\em max-over-time} pooling. From Table 2 we see that more chunks perform better than the {\em max-over-time} pooling. The results indicate that partitioning the sentence into several chunks and summarizing respectively their important semantics can result in better sentence semantic representations.

Overall, the model using 4 chunks (+BCCNN-4) performs best. It obtains three bests out of four sets and it significantly outperforms the model +NNJM on test sets MT06 and MT08. The largest gains can be up to 1.38 BLEU score on MT08. It is interesting that the model with 8 chunks just performs similarly to that with 2 chunks. We speculate that too many chunks may bring some noise. However, it deserves deep investigation. Nevertheless, we can say that the global sentence semantic representation much benefits the target translation prediction.

\section{Related Work}
Our work mainly includes two key issues: one is learning the sentence semantic representation, and the other is applying global sentence-level information to better predict target translations. We will discuss the related work from these two aspects.

On sentence representation learning, many researchers perform this task using the neural network methods.  \namecite{sutskever2014sequence} and \namecite{bahdanau2014neural} applied the recurrent neural networks to encode the source sentence and decode from the source sentence representation. And it takes long time to train their models. \namecite{socher2011parsing,socher2013parsing,socher2013recursive} designed the recursive neural networks for syntactic parsing and sentiment analysis, in which sentence representation is the by-product. \namecite{le2014distributed} used a simple feed-forward neural network to learn sentence and paragraph representations, but one disadvantage is that test sentence representation must be learnt by performing the training process. \namecite{kim2014convolutional}, \namecite{kalchbrenner2014convolutional}, \namecite{hu2014convolutional} and \namecite{zeng2014relation} adopted the convolutional neural networks to learn sentence representations for different classification tasks.

The sentence representations learnt with the above methods are mainly task dependent. For example they are sensitive to sentiment (relation or structure) of the sentence \cite{socher2013recursive,kim2014convolutional,zeng2014relation}. These sentence representations do not encode adequate semantics of the sentence. In contrast, we aim at encoding as much semantics as possible in the sentence representation by designing the bilingually-constrained chunk-based convolutional neural networks.

On applying more information for translation prediction, \namecite{devlin2014fast} developed a neural network joint model to make full use of the source- and target-side local contexts. However, they ignored the global sentence-level information. Recently, some researchers exploited the information beyond the sentence level. For example, \namecite{eidelman2012topic}, \namecite{su2012translation} and \namecite{zhang2014topic} attempted to apply the topic information of the document to distinguish good translation rules from bad ones during SMT decoding. But, their methods require the sentence's document information which is difficult to satisfy in practice. Instead, we just focus on the sentence-level features. We have designed two neural networks: one for sentence semantic representation learning, and the other for target word probability prediction.

\section{Conclusions and Future Work}
In this paper, we have explored the source-side global sentence representations for target translation prediction. We presented a new bilingually-constrained chunk-based convolutional neural network to learn sentence semantic representations. In order to integrate the sentence representation in SMT model, we further applied a feed-forward neural network joint model to better predict translation probability with both local and global information. The extensive experiments have shown that the proposed model can significantly outperform the strong hierarchical phrase-based translation model enriched with broader context features.

At present, the number of chunks in the chunk-based CNN is determined manually. We plan to analyse the sentence structure and design a theoretical way to choose the number of chunks.  We also would like to apply our sentence semantic representations to other tasks such as question answering and paraphrase detection.

\bibliographystyle{named}
\bibliography{senrep4mt}

\end{document}